\documentclass[a4paper]{llncs}

    \usepackage{ifpdf}

    \ifpdf

      \usepackage[pdftex]{graphicx}  

      \usepackage[pdftex]{hyperref}

    \else

      \usepackage[dvips]{graphicx}  

      \newcommand{\href}[2]{#2}

    \fi
\usepackage{amssymb}
\usepackage{graphicx}

\usepackage{url}
\urldef{\mailsa}\path|{vbhalla.du@gmail.com|
\urldef{\mailsb}\path|santanuc@ee.iitd.ernet.in|
\urldef{\mailsc}\path|arihant.jms@gmail.com|    
\newcommand{\keywords}[1]{\par\addvspace\baselineskip
\noindent\keywordname\enspace\ignorespaces#1}

\begin{document}

\mainmatter  

\title{A Novel Hybrid CNN-AIS Visual Pattern Recognition Engine
}

\titlerunning{Lecture Notes in Computer Science: Authors' Instructions}

%
%
\author{Vandna Bhalla
\and Santanu Chaudhury\and Arihant Jain }
\authorrunning{Lecture Notes in Computer Science: Authors' Instructions}

\institute{Indian Institute of Technology \\Delhi, India
\\
\mailsa\\
\mailsb\\
\mailsc\\
\url{http://www.iitd.ac.in}\}}

%
%

\maketitle

\begin{abstract}
Machine learning methods are used today for most recognition problems. Convolutional Neural Networks (CNN) have time and again proved successful for many image processing tasks primarily for their architecture. In this paper we propose to apply CNN to small data sets like for example, personal albums or other similar environs where the size of training  dataset is a limitation, within the framework of a proposed hybrid CNN-AIS model. We use Artificial Immune System Principles to enhance small size of training data set. A layer of Clonal Selection is added to the local filtering and max pooling of CNN Architecture. The proposed Architecture is evaluated using the standard MNIST dataset by limiting the data size and also with a small personal data sample belonging to two different classes. Experimental results show that the proposed hybrid CNN-AIS based recognition engine works well when the size of training data is limited in size.
\end{abstract}

\section{Introduction}

Today all object recognition approaches use machine-learning methods. Larger the dataset better is the performance. Labeled datasets like NORB\cite{1} Caltech-101/256\cite{2}\cite{3} and CIFAR-10/100\cite{4} with tens of thousands of images are in today’s scenario considered small and LabelMe\cite{5} and Image Net \cite{6} with millions of images are preferred.  A simple recognition task also requires datasets of size of the order of tens of thousands of images [14]. It is always assumed that objects in realistic settings show a lot of variability and therefore to learn to recognize them it is essential to have much larger training sets. The many shortcomings of small size data sets have been widely recognized by Pinto\cite{7}. To learn from thousands of objects from millions of images, a model with a large learning capacity with powerful processing is required. 

We present an innovative, adaptive, self-learning, and self-evolving hybrid recognition engine, which works well with small sized training data. The model uses the intelligent information processing mechanism of Artificial Immune System (AIS) and helps Convolutional Neural Network (CNN) generate a robust feature set taking the small set of input training images as seeds. Our model performs visual pattern learning using a heterogeneous combination of supervised CNN and Clonal Selection (CS) principles of AIS. It can be extended to perform classification tasks with limited training data particularly in the context of personal photo collections; where for each training sample different points of view are gathered in parallel using clonal selection.This is very different from populating datasets with artificially generated  training examples\cite{25} by randomly distorting the original training images with randomly picked distortion parameters.

Specific contribution of this paper is as follows: Designed a hybrid Convolutional Neural Network- Artificial Immune System (CNN-AIS) Recognition Engine Architecture designed to work with modest sized training data. This is detailed in Section 3. The model was tested on well-known MNIST digit database and showed remarkable success. The current best rate of 0.3\% on the MINST digit recognition task approaches human performance \cite{8}. But we have got good results with considerably smaller number of training samples. We have also applied this model to a small AIS based classifier and successfully accomplished classification for two categories from a small personal photo collection dataset.

\section{Related Work}
The idea of building a hierarchical structure of features for object detection has deep roots in the computer vision literature\cite{9},\cite{12}. The general structure of the deep convolutional neural network (CNN) was introduced in 1989 by LeCun\cite{13}. His deep convolutional neural network architecture called LeNet is what is still being used today with a lot of consistent improvement to the individual components within the architecture. An important idea of the CNN is that the feature extraction and classifier were unified in a single structure. The parameters of both the classifier and feature detector were trained globally and supervised by back propagation. After the last stage of feature detection a few fully connected layers were added to perform classification. The model was proposed for handwritten digit recognition and achieved a very high success rate on MINST dataset\cite{14}. But it demands substantial amount of labeled data for training (60,000 for MINST). Also the size of input is very small(28x28) with no background clutter, illumination change etc which is an integral part of normal pictures/images. Infact for most realistic vision applications this is not the case. For instance Ranzato et al \cite{15} trained a large CNN for object detection (Caltech 101 dataset) but obtained poor results though it achieved perfect classification on the training set. The weak generalization power of CNN when the number of training data is small and the number of free parameter is large is a case of over fitting or overparametrization. Other biologically inspired models like HMAX \cite{16} use hardwired filter and use hard Max functions to compute the responses in the pooling layer. The problem was that it was unable to adapt to different problem settings.

Successful algorithms have been built on top of handcrafted gradient response features such as SIFT and histograms of oriented gradients (HOG). These are fixed features and are unable to adjust to model the intricacies of a problem. The success of object recognition algorithm to a large extent depends on features detected. The features should have the most distinct characteristics among different classes while retaining invariant characteristics within a class. Traditional hand designed feature extraction is laborious  and moreover cannot process raw images while the automatic extraction mechanism can fetch features directly. The multiple processing  layers of machine learning systems  extract more abstract, invariant features of data and have higher classification accuracy than the traditional shallower classifiers. These deep architectures have shown promising performances in image \cite{14} language \cite{18} and speech\cite{19}.
In  \cite{20}, \cite{21} supervised classifiers such as CNNs, MLPs, SVMs and K-nearest Neighbors are combined in a ‘Mixture of Experts” approach where the output of parallel classifiers is used to produce the final result. CNNs though efficient at learning invariant features from images, do not always produce optimal classification and SVMs with their fixed kernel function are unable to learn complicated invariances. Our approach is different as we propose a single architecture for training and testing using CNN and AIS principles.

\section{Convolutional Neural Network- Artificial Immune System (CNN-AIS) Model}

Our proposed architecture integrates Clonal Selection (CS) principles from Artificial Immune System(AIS) with Convolutional Neural Networks(CNN). We will briefly introduce the Artificial Immune System (AIS) theory and the basic CNN structure that we have used in the subsequent sections. Then the hybrid CNN-AIS trainable recognition engine is presented followed by results and analysis of its merits.

\subsubsection{Artificial Immune Systems (AIS):}
AIS use Clonal Selection and Negative Selection imitating the biological immune system. The main task of the immune system is to defend the organism against pathogens. In the human body the B-cells with different receptor shapes try to bind to antigens. The best fit cells proliferate and produce clones which mutate at very high rates. The process is repeated and it is likely that a better B-cell (better solution) might emerge. This is called Clonal Selection. These clones have mutated from the original cell at a rate inversely proportional to the match strength. Two main concepts are particularly relevant for our framework.
Generation of Diversity: The B cells produce antibodies for specific antigens. Each B cell makes a specific antibody, which is expressed from the genes in its gene library. The gene library does not contain genes that define antibodies for every possible antigen. Gene fragments in the gene library randomly combine and recombine and produce a huge diverse range of antibodies

This helps the immune system to make the precise antibody for an antigen it may never have encountered previously.                                        
Avidity: Refers to the accrued strength of various diverse affinities of individual binding interaction. Avidity (functional affinity) is the collective strength of multiple affinities of an antigen with various antibodies. 
Based on this biological process, quite a few Artificial Immune System (AIS) have been developed in the past, [22] and [23]. Castro developed the Clonal Selection Algorithm (CLONALG) [24] on the basis of Clonal Selection theory of the immune system. It was proved that it can perform pattern recognition. The CLONALG algorithm can be described as follows:
1. Randomly initialize a population of individual (M);
2. For each pattern of P, present it to the population M and determine its affinity with each element of the population M;
3. Select n of the best highest affinity elements of M and generate copies/clones of these individuals proportionally to their affinity with the antigen which is the pattern P. The higher the affinity, the higher the number of clones, and vice-versa;
4. Mutate all these copies with a rate proportional to their affinity with the input pattern: the higher the affinity, the smaller the mutation rate;
5. Add these mutated individuals to the population M and reselect m of these maturated individuals to be kept as memories of the systems;
6. Repeat steps 2 to 5 until a certain criterion is met.

\subsubsection{Convolutional Neural Network(CNN)}
A Convolutional Neural Network \cite{25}\cite{26} is a multilayer feed forward artificial neural network with a deep supervised learning architecture. The ordered   architectures of MLPs progressively learn the higher level features with the last layer giving classification. Two operations of convolutional filtering and down sampling alternate to learn the features from the raw images and constitute the feature map layers.
The weights are trained by a back propagation algorithm using gradient descent approaches for minimizing the training error. We have used Stochastic Gradient Approach as it avoids being stuck in poor local minima which is highly likely due to the non linear nature of the error surface. A simplified CNN was presented in \cite{27} which we have used for our work instead of using the rather complicated LeNet-5\cite{28}. The model has five layers.

\subsubsection{CNN-AIS Model}
Modern architecture trains learning features across hidden layers starting from low level details up to high level details. The architecture of our hybrid CNN-AIS model was designed by adding an additional layer of Artificial Immune System (AIS) based Clonal Selection (CS) in the traditional Convolutional Neural Network (CNN) structure, Fig.1. The model is explained layer wise.

\begin{figure}
\hspace{-0.5cm}
\includegraphics[height=8.2cm]{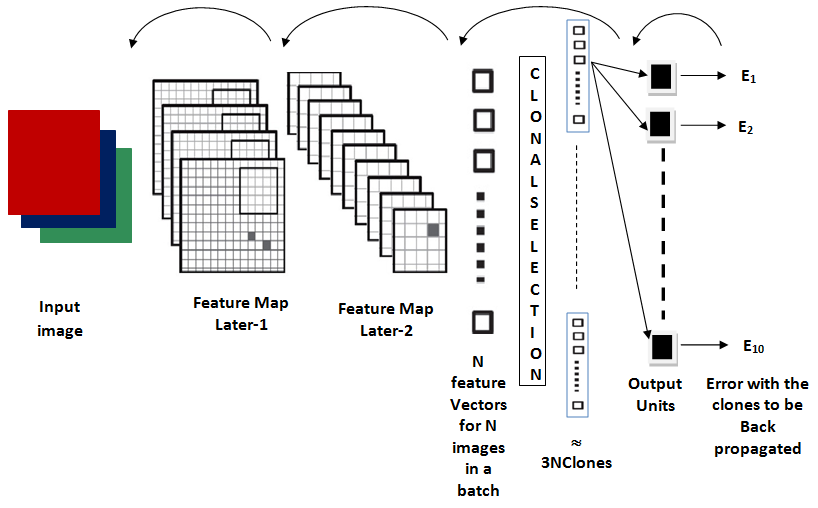}
\caption{CNN-AIS Hybrid Model}
\label{fig:example}
\end{figure} 

Convolutional Layer: A 2D filtering between input images n, and a matrix of kernels/weights K produces the output I where $I_{k}$ = $\Sigma_{ i, j, k}$ є M ($n_i$ * $K_{j}$) where M is a table of input output relationships. The kernel responses from the inputs connected to the same output are linearly combined. As with MLPs a scaled hyperbolic tangent function is applied to every I.

Sub sampling Layer: Small invariances to translation and distortion is accomplished with the Max-Pooling operation. This is for faster convergence and improves generalization as well.

Fully Connected Layer - I: The input to this layer is a set of feature maps from the lower layer which are combined into a 1-dimensional feature vector and subsequently passed through an activation function

\textbf{Clonal selection Layer}: This is the new additional layer that we propose in our architecture
and it is the second last layer. This layer receives its input from the fully connected layer-I in the form of 1-D feature vector for all the images (n) in the current running batch. Each feature vector in the Feature set undergoes Cloning, Mutation and Crossover according to the rules of Clonal Selection  to generate additional features that satisfy the minimum threshold criteria and resemble the particular class.
The number of clones is calculated by
\\CNum= $\eta$η x affinity (Feature Vector1, Feature Vector2) 	... (i), where $\eta$η is the cloning constant.
Higher the affinity of match the greater the clone stimulus gets, the more the cloning number is. On the contrary, the number is less, which is consistent with biological immune response mechanism. 
Mutation frequency is defined as Rate, which is calculated by 

Rate = $\alpha$ {1/ affinity (Feature Vector1, Feature Vector2)} … (ii)
Where $\alpha$α is mutation constant. In accordance with (ii), the higher the affinity of match, the smaller the clone stimulus gets, the lower the mutation frequency is. On the contrary, the mutation frequency is higher.
Hence from n initial feature sets we now have (n  x CNum) feature sets. These newly generated feature vectors are grouped into batches and individually fed to the output layer and the subsequent error is backpropagated  to train the kernels of the CNN. Hence from the seeds of a few representative images of each class a bigger set is evolved using Clonal Selection principles of Artificial Immune System. End of training phase yields a set of representative features, which we call antibodies, from each class of size much larger than the original dataset and a trained CNN. Though we start with random values of feature sets(antibodies) for each class but eventually they converge to their  optimal values.  

Output Layer(Fully Connected Layer-II): This layer has one output neuron per class label and acts as linear classifier operating on the 1-dimensional feature vector set computed from the CS layer

\section{Result}
We performed tests on MNIST dataset. Fig.2 is a plot of comparison of error versus training data size for both CNN and CNN +AIS hybrid. When available data is less then CNN+AIS model performs better giving lesser error.It is evident that CNN-AIS error rates are much lower for small data size. Hence AIS helps in training CNN better when training data is scarce. Fig.3 shows a plot of error versus the number of epochs for our hybrid model for standard data size. As the number of epochs increases the error rate decreases and becomes constant after 15 epochs.

\begin{figure}
\centering
\centering
\includegraphics[height=7cm]{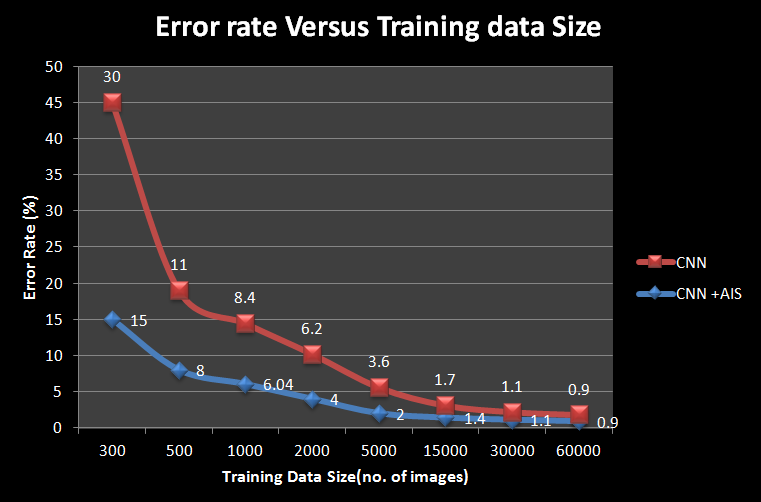}
\caption{Error versus training data size }
\label{fig:example}
\end{figure}

\begin{figure}
\centering
\includegraphics[height=7cm]{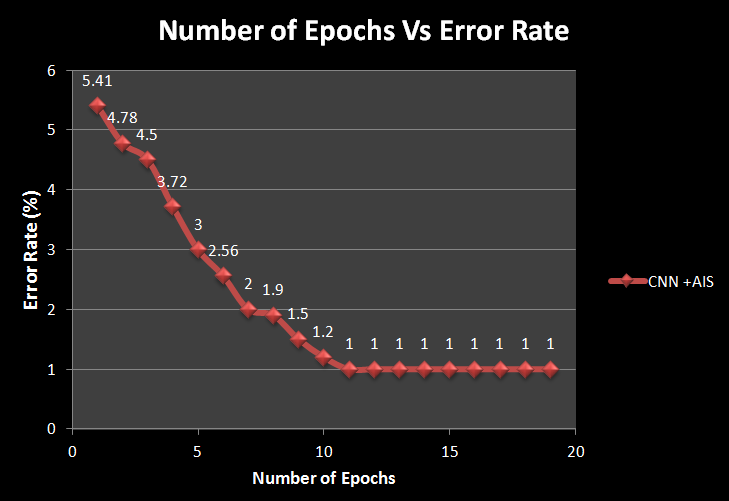}
\caption{Plot of error rate w.r.t no. of epochs  }
\label{fig:example}
\end{figure}

\section{Application: Personal Photo Album}
The CNN-AIS generates a robust and diverse pool of feature vectors and a trained CNN for any class. We tested this model for a personal collection of photos for two classes Picnic (A) and Conference (N). For every testing image (the antigen), the trained CNN-AIS model computes the feature vector and compares this with the feature set pool of that class. If the number of matches of the test image with the various feature sets of that class and the combined affinities exceed the threshold then the testing image is placed in that class. These emulate the antibodies in a human body recognizing an antigen. The model is shown in Fig 4.
\begin{figure}
\centering
\includegraphics[height=8.2cm]{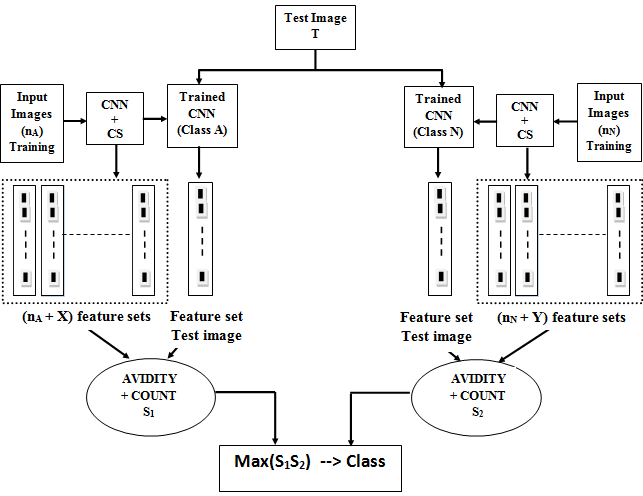}
\caption{Application: CNN-AIS Model Used for Personal Photo Classification}
\label{fig:example}
\end{figure}

\begin{figure}
\centering
\includegraphics[height=2.6cm]{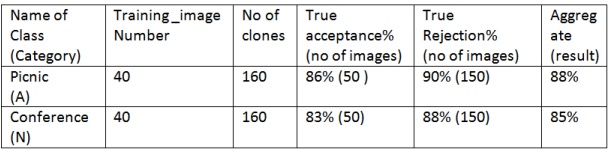}
\caption{Application Analysis on Personal Photo Album}
\label{fig:example}
\end{figure}

\begin{figure}
\centering
\includegraphics[height=4cm]{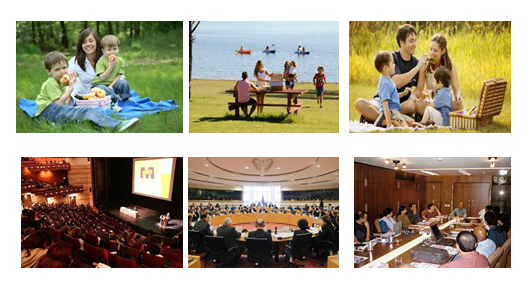}
\caption{Sample images from Application dataset}
\label{fig:example}
\end{figure}
The trained CNN learns the features of the test image, the antigen. A two phase testing mechanism is used for classification. The first phase matches the test image feature with the 3N feature sets (antibodies) of each of the classes. The total number of antibodies lying above the threshold for matching is counted (C) for each class.  All classes providing a minimum number of C are qualified for phase 2 testing. The second phase calculates Avidity for each class which is the mean strength of multiple affinities of all qualified antibodies in C with the testing image antigen. It is calculated by taking the mean of individual scores (calculated using inner product measure) of matching of test image with each antibody above the threshold for each qualified class. This score is labeled avidity. The class is eventually decided on the basis of the combined scores S=(Count+Avidity). The test images are different from training images and may belong to different individuals. The model can be extended to generate a new class. If during testing no suitable match happens then the test image can initialize a new class/ antibody set with its features set. Mutation generates the new relevant population. The experimental results are summarized in Fig.5. The sample dataset used for our experiments is shown in Fig.6. Despite the diversity in the dataset and the small size of training data set, our model gives good results.

\section{Conclusion}
The AIS layer shows a marked improvement in recognition when training data is limited. The proposed model can be extended to be used as a classifier for personal photo albums as explained by the example application. The results are very encouraging. A new class can be added to the existing set of classes dynamically replicating the behavioral aspects of self-learning and self evolving of human immune system. The results show the efficacy of our model. The proposed model is able to capture diversity that is inherent in personal photo collections unlike CNN by itself.


\begin{thebibliography}{4}

\bibitem{1} Y. LeCun, F.J. Huang, and L. Bottou. Learning methods for generic object recognition with invariance to pose and lighting. In Computer Vision and Pattern Recognition, 2004. CVPR 2004. Proceedings of the2004 IEEE Computer Society Conference on, volume 2, pages II–97. IEEE, 2004
\bibitem{2} L. Fei-Fei, R. Fergus, and P. Perona. Learning generative visual models from few training examples: Anincremental bayesian approach tested on 101 object categories. Computer Vision and Image Understanding, 106(1):59–70, 2007.
\bibitem{3} G. Griffin, A. Holub, and P. Perona. Caltech-256 object category dataset. Technical Report 7694, California Institute of Technology, 2007. URL http://authors.library.caltech.edu/7694.
\bibitem{4} A. Krizhevsky. Learning multiple layers of features from tiny images. Master’s thesis, Department ofComputer Science, University of Toronto, 2009.
\bibitem{5}B.C. Russell, A. Torralba, K.P. Murphy, and W.T. Freeman. Labelme: a database and web-based tool for image annotation. International journal of computer vision, 77(1):157–173, 2008.
\bibitem{6} J. Deng, W. Dong, R. Socher, L.-J. Li, K. Li, and L. Fei-Fei. ImageNet: A Large-Scale Hierarchical Image Database. In CVPR09, 2009.
\bibitem{7} N. Pinto, D.D. Cox, and J.J. DiCarlo. Why is real-world visual object recognition hard? PLoS computational biology, 4(1):e27, 2008.
\bibitem{8}D. Cire¸san, U. Meier, and J. Schmidhuber. Multi-column deep neural networks for image classification. Arxiv preprint arXiv:1202.2745, 2012.
\bibitem{9} Fukushima. Neocognitron: A self-organizing neural network model for a mechanismof pattern recognition unaffected by shift in position. Biological Cybernetics, 36(4):193–202, 1980

\bibitem{12} M. Ranzato, F. J. Huang, Y. Boureau, and Y. LeCun. Unsupervised learning of invariant feature hierarchies with applications to object recognition. In Proc. IEEE Conference Computer Vision and Pattern Recognition, 2007.
\bibitem{13}Y. LeCun, L. D. Jackel, B. Boser, J. S. Denker, H. P. Graf, I. Guyon, D. Henderson, R. E. Howard and W. Hubbard: Handwritten Digit Recognition: Applications of Neural Net Chips and Automatic Learning, IEEE Communication, 41-46, invited paper, November 1989,
\bibitem{14} Alex Krizhevsky and Ilya Sutskever and Geoffrey E. Hinton: Imagenet classification with deep convolutional neural networks, Advances in Neural Information Processing Systems,2012
\bibitem{15} M. Ranzato, F. J. Huang, Y. Boureau, and Y. LeCun. Unsupervised learning of invariant feature hierarchies with applications to object recognition. In Proc. IEEE Conference Computer Vision and Pattern Recognition, 2007.
\bibitem{16}T. Serre, L.Wolf, S. Bileschi, M. Riesenhuber, and T. Poggio. Robust object recognition with cortex-like mechanisms. IEEE Trans. PAMI, 29(3):411–426, 2007

\bibitem{18} D. Yu, L. Deng, and S. Wang,“ Learning in the deep structured conditional random fields,”in Proc. Neural Inf. Process. Syst. Workshop, Vancouver, BC, Canada, Dec. 2009, pp. 1–8.
\bibitem{19}A. R. Mohamed, T. N. Sainath, and G. Dahl,“ Deep belief networks using discriminative features for phone recognition,”in Proc. Acoust. Speech Signal Process. (ICASSP), Prague, Czech Republic, 2011, pp. 506–5063
\bibitem{20}S. Abdelazeem, “A greedy approach for building classification cascades,” in Proc. of the 7th ICMLA, 2008,pp. 115–120
\bibitem{21} A. Borji, “Combining heterogeneous classifiers for network intrusion detection,” in Proc. of the 12th Asian Computing Science Conf. (ASIAN), 2007, pp. 254–260.
\bibitem{22} E. Hart and J. Timmis. Application areas of ais: The past, the present and the future. 2008..  
\bibitem{23} G. Dudek. An artificial immune system for classification with local feature selection. IEEE Trans. Evolutionary Computation, 2012
\bibitem{24} Castro, L. N. and Von Zuben, F. J :Learning and Optimization Using the Clonal Selection Principle, Journal Trans. Evol. Comp, June 2002 volume {6},IEEE Press
\bibitem{25} Y. LeCun, L. Bottou, Y. Bengio, P. Haffner, Gradient-based learning applied to document recognition, Proceedings of the IEEE 86 (11) (1998) 2278–2324. 
\bibitem{26} Convolutional Neural Support Vector Machines: Hybrid Visual Pattern Classifiers
for Multi-robot Systems, 2012 11th International Conference on Machine Learning and Applications
\bibitem{26} F. Lauer, C.Y. Suen, G. Bloch, A trainable feature extractor for handwritten digit recognition, Pattern Recognition 40 (6) (June 2007) 1816–1824.
\bibitem{27}	W. Pan, T.D. Bui, C.Y. Suen, Isolated handwritten Farsi numerals recognition using sparse and over-complete representations, in: Proceedings of the International Conference on Document Analysis and Recognition, Barcelona, Spain, July 2009, pp. 586–590.
\bibitem{28}F. Lauer, C.Y. Suen, G. Bloch, A trainable feature extractor for handwritten digit recognition, Pattern Recognition 40 (6) (June 2007) 1816–1824


\end{thebibliography}
\end{document}